\documentclass[11pt,a4paper,twoside]{article}

\usepackage{fullpage}
\usepackage[utf8]{inputenc}
\usepackage[T1]{fontenc}
\usepackage{float}

\usepackage{amsmath, amsthm, amssymb, amsfonts}
\usepackage{thmtools, thm-restate}
\usepackage{bm}
\usepackage{mathtools}
\usepackage{graphicx}
\usepackage{caption}
\usepackage{subcaption}
\usepackage{color}
\usepackage{xcolor}

\graphicspath{{figures/}}

\DeclareFontFamily{U}{matha}{\hyphenchar\font45}
\DeclareFontShape{U}{matha}{m}{n}{
      <5> <6> <7> <8> <9> <10> gen * matha
      <10.95> matha10 <12> <14.4> <17.28> <20.74> <24.88> matha12
      }{}
\DeclareSymbolFont{matha}{U}{matha}{m}{n}

\DeclareFontFamily{U}{mathx}{\hyphenchar\font45}
\DeclareFontShape{U}{mathx}{m}{n}{
      <5> <6> <7> <8> <9> <10>
      <10.95> <12> <14.4> <17.28> <20.74> <24.88>
      mathx10
      }{}
\DeclareSymbolFont{mathx}{U}{mathx}{m}{n}

\DeclareMathDelimiter{\vvvert}{0}{matha}{"7E}{mathx}{"17}

\usepackage{stmaryrd}

\usepackage{dsfont}

\usepackage{calc}

\usepackage{algorithm,algorithmicx,algpseudocode}

\usepackage{hyperref}
\hypersetup{
 pdfauthor={},
 pdftitle={},
 pdfsubject={},
 pdfkeywords={}
}
\usepackage{cleveref}
\usepackage{ulem}

\usepackage{datatool}
\DTLsetseparator{;}
\DTLloadrawdb[noheader, keys={thekey,thevalue}]{runsinfo}{runsinfo.csv}
\newcommand{\perplexityinsert}[1]{\DTLgetvalueforkey{\datavalue}{thevalue}{runsinfo}{thekey}{#1}\datavalue}

\newtheorem{theorem}{Theorem}[section]

\theoremstyle{definition}

\newtheorem{remark}[theorem]{Remark}

\numberwithin{equation}{section}

\newcommand{\R}{\ensuremath{\mathbb{R}}}

\newcommand{\cC}{\ensuremath{\mathcal{C}}}

\newcommand{\cF}{\ensuremath{\mathcal{F}}}

\newcommand{\cS}{\ensuremath{\mathcal{S}}}
\newcommand{\cT}{\ensuremath{\mathcal{T}}}

\newcommand{\cX}{\ensuremath{\mathcal{X}}}

\newcommand{\bB}{\ensuremath{\mathbb{B}}}

\newcommand{\bP}{\ensuremath{\mathbb{P}}}

\newcommand{\bR}{\ensuremath{\mathbb{R}}}

\newcommand{\G}{\ensuremath{\textsf{G}}} 
\newcommand{\E}{\ensuremath{\textsf{E}}} 
\newcommand{\V}{\ensuremath{\textsf{V}}} 

\newcommand{\vv}{\ensuremath{\textsf{v}}}
\newcommand{\ww}{\ensuremath{\textsf{w}}}
\newcommand{\ee}{\ensuremath{\textsf{e}}}
\newcommand{\cc}{\texttt{c}}

\newcommand{\green}{\ensuremath{\texttt{green}}}
\newcommand{\yellow}{\ensuremath{\texttt{orange}}}
\newcommand{\red}{\ensuremath{\texttt{red}}}
\newcommand{\dark}{\ensuremath{\texttt{dark-red}}}

\newcommand{\<}{\langle}
\renewcommand{\>}{\rangle}

\def\({\left (}
\def\){\right )}

\newcommand{\vspan}{\operatorname{span}}
\newcommand{\argmin}{\operatorname*{argmin}}

\newcommand{\obs}{\ensuremath{\text{obs}}}

\newcommand{\test}{\text{test}}
\newcommand{\train}{\text{train}}

\newcommand{\dd}{\ensuremath{\mathrm d}}
\newcommand{\dr}{\ensuremath{\mathrm dr}}

\begin{document}
\begin{filecontents*}{runsinfo.csv}
filtergraph;True
shuffletimes;False
screenshotperiod;15
screenshotwidth;1253
screenshotheight;1253
zoom;13
imagex;800
imagey;1000
numstationstest;10
stationstest;OPERA, BONAP, ELYS, PA07, PA13, PA18, BASCH, PA12, CELES, HAUS
percentagetraintime;80
numstations;13
numberoftraintimes;374
numnodes;12963
numedges;25476
numnodesfilter;10116
numedgesfilter;18713
MinDistNodeStation;165
green;(99, 214, 104)
yellow;(255, 151, 77)
red;(242, 60, 50)
darkred;(129, 31, 31)
kneighbours;10
hiddenlayers;2
activation;logistic
solver;adam
neurons;20
average;Average in space
krigging;Krigging
avgkrigging;Global Average Krigging
lm;Graph Linear Regression
lmextra;Graph Temp Wind Linear Regression
nn;Graph Temp Wind NeuralNetwork
ensemble;Ensemble
\end{filecontents*}

\title{State estimation of urban air pollution\\
with statistical, physical, and super-learning graph models}
\author{Matthieu Dolbeault, Olga Mula, and Agustín Somacal
}
\date{}
\maketitle

\begin{abstract}
We consider the problem of real-time reconstruction of urban air pollution maps. The task is challenging due to the heterogeneous sources of available data, the scarcity of direct measurements, the presence of noise, and the large surfaces that need to be considered. In this work, we introduce different reconstruction methods based on posing the problem on city graphs. Our strategies can be classified as fully data-driven, physics-driven, or hybrid, and we combine them with super-learning models. The performance of the methods is tested in the case of the inner city of Paris, France.
\end{abstract}

\section{Introduction}

\subsection{Background and motivation}
Data-driven estimations are becoming increasingly relevant and widespread as the volume and heterogeneity of available data increases. A fundamental challenge is to build numerical methods for which one can estimate how optimally they exploit the given information.
The present paper addresses some essential computational aspects connected to this question. More specifically, our goal is to reconstruct a state $u$ of a physical process, for which we have at hand very heterogeneous sources of data coming from direct partial observations of $u$, from quantities related to $u$, and from the knowledge that the physics can be modelled by a Partial Differential Equations (PDE).

Assume that $u$ belongs to some Banach space $U$ of potentially infinite dimension, with associated norm $\Vert \cdot \Vert_U$, and that all the available information is given by an element $x_u$ from some abstract metric space $\cX$. Our goal is thus to build a mapping $A:\cX\to U$ such that $A(x_u)$ approximates $u$ at best, in the sense that the approximation error
\begin{equation}
e(A,u)=\Vert u - A(x_u)\Vert_U
\label{eq:text}
\end{equation}
is as small as possible,
for any configuration $(u,x_u)$ of the system.
In practice, finding the optimal map $A$ is not feasible, and various suboptimal reconstruction techniques have been proposed,
each of them having its own virtues and drawbacks: statistical approaches such as BLUE \cite{BLUE} and kriging \cite{kriging}, model order reduction of parametric PDEs \cite{ROM, Mula2023}, or more recently approximations by neural networks and machine learning strategies \cite{NN-inv-1, NN-inv-2}. Since all of these strategies are sub-optimal, and each one is based on different a priori assumptions, one should not make the methods compete against each other, but rather collaborate with each other to enhance their respective strengths. This leads naturally to explore approaches based on ensemble super-learning \cite{Breiman1996, LPH2007, PL2010} as we consider in the present work.

\subsection{Urban air pollution modeling} 

There are numerous applications in which one is confronted with the above state estimation problem. As a guiding example, we consider in this paper the real-time reconstruction of urban pollution fields. Beyond the relevance of such a task to limit environmental and health risks in the city, pollution state estimation is an excellent example where collaborative, super-learning methods are required. This is because the problem accumulates several difficulties that make the reconstruction challenging for most common reconstruction methods. Among the issues, we may mention the following:

\begin{itemize}
\item \emph{Scarcity of pollution measurements:} The amount of reliable sensor devices measuring pollutant concentrations is often limited, and the measurements are usually taken at fixed locations. As a result, reconstruction methods based solely on these measurements lack spatial resolution, and exhibit huge uncertainties in regions without sensors.
\item \emph{Heterogeneous data:} In addition to the pollutant measurements, other sources of relevant information are available such as traffic estimations in each street, wind speed, topography, temperature, etc. However, it is not obvious how to meaningfully combine this data to enhance the estimation. Some attempts have been tried in \cite{criado_data_2023} through the use of Gradient Boosting Machines and Universal Krigging, with positive results for the estimation of PM10 particles in the city of Barcelona.
\item \emph{Lack of training data:} Even when incorporating other sources of information, the available data may be insufficient, noisy or hardly correlated to the pollution levels we wish to caracterize. Purely data-driven models greatly suffer from these impediments in their training phase.
\item \emph{Complexity of the physical problem:} The equations governing the dispersion of pollutants in the atmosphere are nonlinear, with turbulent effects at the street scale, thus imposing a fine spatial resolution, at least near the sensor stations \cite{criado_data_2023}. On the other hand, the computational domain is of the size of a city, making it prohibitively expensive to solve a full model like 3D Navier-Stokes equations.
\item  \emph{Parameter calibration:} Reduced models use effective parameters, which account for large-scale averages of local effects, in order to alleviate the requirements on the resolution. However these parameters must be calibrated based on the available data or preliminary simulations, which is a hard task given the above issues.
\end{itemize}
The above obstructions advocate for collaborative strategies combining physics-driven and data-driven approaches such as the one that we develop in this paper. A similar idea has been explored in \cite{niu_novel_2016}, but for forecasting temporal series of pollutant, instead of performing state estimation on a large spatial domain.

It should be noted that the limited number of reliable measurements will still pose problems for validating and assessing the quality of each model, which is a crucial part in collaborative strategies.
We will mitigate this defect by operating multiple leave-one out cross validations, which preserve as much data as possible for the training part of each model, while testing them on many instances.

\subsection{Contributions and layout of the paper}
Our main contributions are:
\begin{enumerate}
\item the construction of numerous physics-based and data-driven models for state estimation;
\item the construction of a very general ensemble super-learning method combining the above models;
\item its application to the task of recovering urban pollution maps at a city scale, together with a comparison of its constitutive submodels;
\item and the development of a routine extracting car emissions in each street from traffic maps. Moreover, we have created a dataset comprised of processed traffic data from Google Maps screenshots, which can be used for future research.
\end{enumerate}

In our numerical experiments, we work with the inner city of Paris, which covers a surface of about $140\,\textrm{km}^2$. The pollutant we consider is $\textrm{NO}_2$,
which is monitored for its respiratory effects,
while being mainly produced by vehicle emissions. We use concentration measurements from Airparif sensors\footnote{We extracted data from the Airparif database, which can be found at \url{https://data-airparif-asso.opendata.arcgis.com}}, and real-time traffic data from Google Maps\footnote{The permission to use Google Maps data for non-profit research
is stated here: \url{https://about.google/brand-resource-center/products-and-services/geo-guidelines/\#google-maps}}. Compared to previous contributions and other existing reconstruction methods (see, e.g., \cite{MSM2009, TMPPB2013}),
the use of such online traffic data is rather novel. It gives a rough estimation of the spatial density of street traffic, benefits from a very fine spatial resolution, and can be freely updated as frequently as desired, in contrast to many existing approaches which only use time averages of traffic data.

Another distinctive aspect of our approach is the representation of the city by a graph, where nodes and edges correspond to crossroads and street segments, instead of considering an open subset of $\bR^2$ or $\bR^3$ as the spatial domain. This description immediately includes geometric specificities of the agglomeration under study, such as the orientation of each street or the configuration of each neighborhood. It is a natural framework for taking into account pollutant emissions caused by traffic, which are located on the graph. Moreover, it is in adequacy with our goal of estimating local variations in the concentration of pollutants close to the ground, since the streets are isolated from each other at this height.

Physical models can be solved on such domains thanks to the theory of \emph{quantum graphs}, that is, metric graphs endowed with a differential operator acting on functions defined on the graph (see \cite{BK2013} for details and references). The metric graph structure leads to the definition of suitable and natural function spaces to pose the problem. Of course, several physical models of different complexity could be considered. In this paper, we work with simple elliptic operators, but the model could be refined by considering, for instance, advection-diffusion operators.
\newline

The rest of the paper is organized as follows. In Section \ref{sec:problem-setting} 
we present our guiding numerical example of the Parisian area, and the available data. Section \ref{sec:concrete-methods} explains the different reconstruction methods we have used for our numerical experiments, including ensemble super-learning methods combining the previous ones. By construction, the super-learner has higher approximation power than each individual model. Section \ref{sec:rigorous} discusses how to theoretically quantify performance and optimality of the numerical algorithms, and why leave-one-out is a good way to estimate this performance in practice. We summarize our numerical experiments and provide some illustrations in Section \ref{sec:numerical-tests}. 
Finally, Appendix~\ref{sec:metric-graphs} details the mathematical setting for the problem of pollution state estimation on graphs.

\section{Available data and pre-processing}
\label{sec:problem-setting}

\subsection{Pollution sensors}
\label{sec:pollution-sensors}

\begin{figure}
    \centering
    \includegraphics[width=0.6\textwidth, trim=0 1cm 0 2cm]{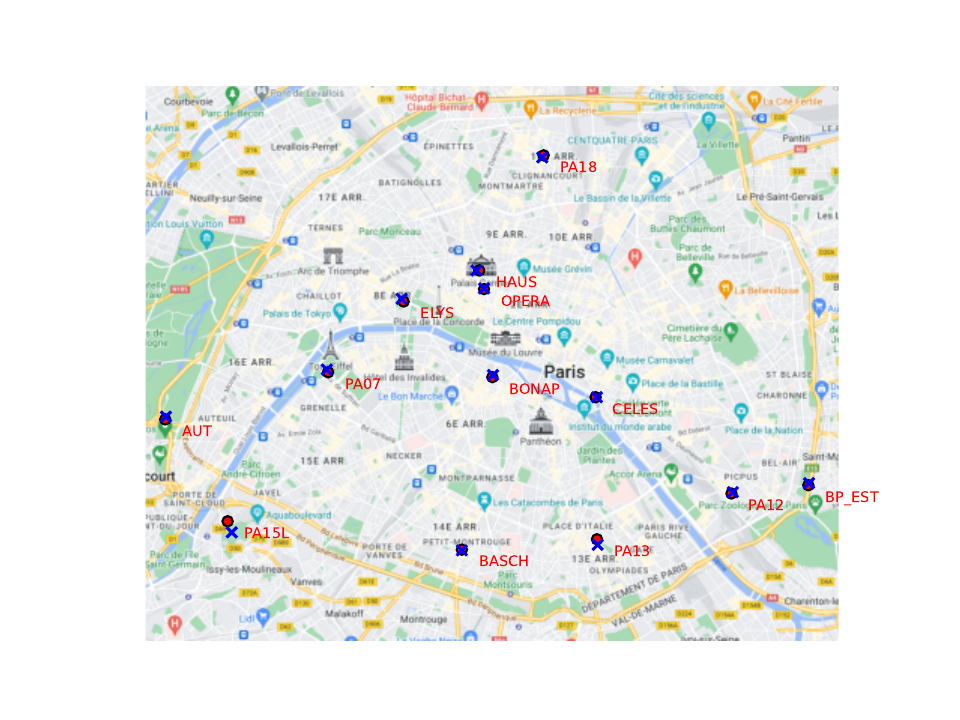}
    \caption{Cropped Google Map screenshot of Paris and the $m=13$ available stations in the study: red dots represent the projection of the station locations to the nearest vertex in the graph of streets, while the blue crosses correspond to the exact position of the station.}
     \label{fig:paris-stations}
\end{figure}

The main information we use consists of direct measurements of the $\textrm{NO}_2$ concentration field~$u$ at Airparif sensor stations. There are $m=\perplexityinsert{numstations}$ such stations, which are placed at fixed locations
\[
r^\obs \coloneqq\{r_1^\obs, \dots, r_m^\obs\} \in (\bR^2)^m,
\]
see Figure~\ref{fig:paris-stations}. Each of the stations provides hourly averages of the concentration of nitrogen dioxide, in $\mu \text{g/m}^3$, of the form
\[
z_i = u(r_i^\obs) + \eta_i, \quad i=1,\dots, m,
\]
where $\eta_i$ is some noise with nominal relative error $|\eta_i|/u(r_i^\obs) \leq 15\%$.

Note that, in principle, the equations governing the dispersion of pollutants are time dependent. However, as we only have measurements every hour, we opt for a static model, where the state at a given time is computed based on the data available at this time only. This essentially amounts to assuming that the emissions vary slowly over time, and that the system reaches an equilibrium state in less than one hour.

\subsection{Meteorological conditions}

Wind, as well as stratification effects in the atmosphere due to variations in temperature, play a major role in the dispersion of pollutants \cite{cai2022effects}. Moreover,
the chemical equilibrium between $\textrm{NO}$ and $\textrm{NO}_2$ depends on the cloud cover \cite{li2020quantifying}. Therefore, we collect the temperature $\theta\in \bR$ and the wind speed $w\in \bR^2$ at every hour from a weather archive \footnote{See \url{https://www.windguru.cz}. For the wind, we combined the absolute wind speed with the wind direction to obtain a vector in $\bR^2$.}.
These two quantities are treated as global, that is, they are assumed to be constant over the spatial domain.

\subsection{Traffic}

\begin{figure}
          \centering
          \includegraphics[width=0.6\textwidth]{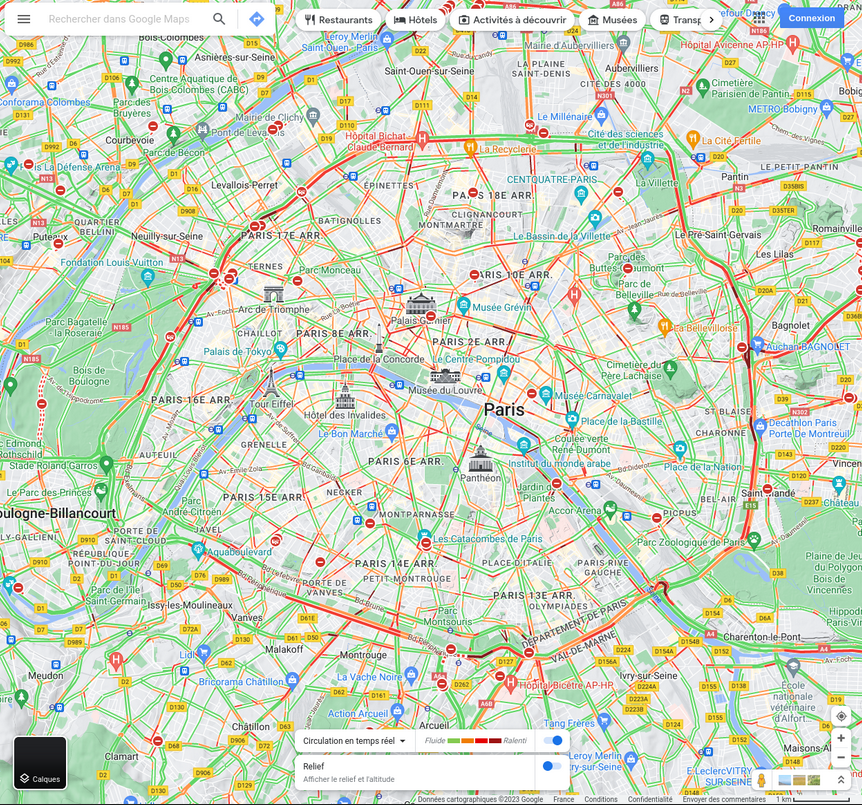}
          \caption{Raw data from Google Maps: the image contains the city with its main landmarks, and some streets are highlighted with one of the four colors corresponding to traffic.}
          \label{fig:traffic-raw-data}
 \end{figure}

Car traffic is responsible for more than half the emissions of $\textrm{NO}_2$ in urban environments \cite{kurtenbach2012primary}.
There is an increasing number of available sources that give access to traffic data. In our case, we work with traffic information extracted from Google Maps. We have designed a script using the Python library Selenium to automatically take screenshots of Paris
every $\perplexityinsert{screenshotperiod}$ minutes over an area of $\perplexityinsert{screenshotwidth} \times \perplexityinsert{screenshotheight}$ pixels with zoom level $\perplexityinsert{zoom}$. An example of resulting \textit{raw image} can be seen in Figure~\ref{fig:traffic-raw-data}. 
Note that city landmarks could not
be removed before taking the screenshot, nor even by substracting a background image, since each screenshot has slight color variations, rendering this approach impractical. Another issue is the absence of traffic data in the smallest
streets of the city. In addition, linking it to the pollution field requires some calibration.

On the one hand, this kind of information is very rich because of its availability in real time, and its spatial coverage of the whole city at high resolution. On the other hand, it is very partial: it comes in the form of four colors, each representing a certain traffic intensity, which gives a qualitative estimate of the number of cars in each street: a street marked in red is, for instance, more congested than one marked in green.
One main goal in our work is to examine the potential of incorporating such low-quality information for state estimation tasks.

\label{sec:paris-graph}

\begin{figure}[htb]
\centering
\includegraphics[width=0.6\textwidth, trim=0 5cm 0 5cm]{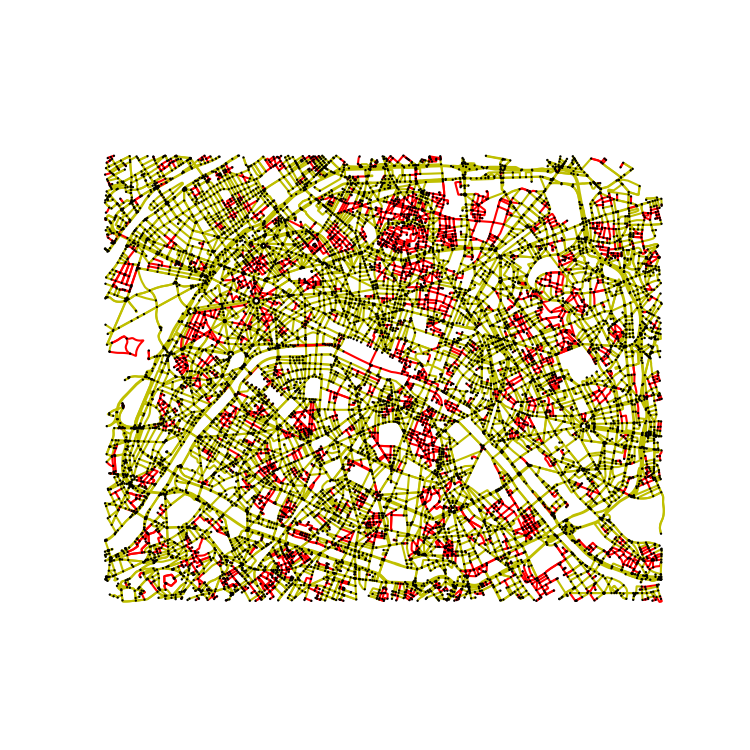}
\caption{The metric graph downloaded from Open Street Maps, with the edges that never had Google Traffic activation in red, and the edges remaining after filtration in yellow}
\label{fig:paris-graph}
 \end{figure}
 
\subsection{Graph of Paris}
 
In order to locate our sources and to express the spatial dependence of a state, we consider a graph domain. 
We use a metric graph $\G=(\V,\E)$ provided by Open Street Maps, together with the Python library osmnx. For the mathematical definition of metric graphs and their associated function spaces, we refer to Appendix \ref{sec:metric-graphs}. The graph $\G$ covers the whole inner ring of the city, as shown in Figure~\ref{fig:paris-graph}. The full graph has $|\V|=\perplexityinsert{numnodes}$ vertices and $|\E|=\perplexityinsert{numedges}$ edges, but we restrict it to the biggest connected component of the subgraph that remains after filtering out all the edges which have never been colored with traffic information. After this operation, our actual graph has $|\V|=\perplexityinsert{numnodesfilter}$ vertices and $|\E|=\perplexityinsert{numedgesfilter}$ edges. The street network is relatively dense, most nodes having 3 to 6 edges.

The vertices $\vv \in \V$ come with precise geographical coordinates. In the following, we assume that the graph is embedded in the two-dimensional plane, and do not take altitude into account. Each edge $\ee\in \E$ is a street or a portion of it, and we have access to its length $\ell_\ee$ as well as its shape, number of lanes and speed limit. The information is so detailed that the streets are represented by paths that are not necessarily straight lines. However, in the following, 
we will work under this slightly simplified setting.

The location of the sensor stations does not exactly match with vertices of the graph. We therefore project their position $r^\obs$ onto the nearest vertex,
which yields observational nodes
\[
\vv_i^\obs\coloneqq \argmin_{\vv\in \V} |r_i^\obs-\vv|,\quad i=1,\dots,m,
\]
As Figure \ref{fig:paris-stations} illustrates, the projected locations are very close to the exact locations, with a maximal discrepancy of $\perplexityinsert{MinDistNodeStation}$m, to be compared with the width of the domain, of about $12$km. As a consequence, we will assume that the observations $z_i$ correspond to the values $u(\vv_i^\obs)$, up to a slight increase in the measurement errors $|\eta_i|$.

\subsection{Pre-processing of traffic data}
\label{sec:traffic-preprocess}

We also map the traffic information onto pollutant emissions on the graph edges,
by implementing the following pipeline:
\begin{itemize}
\item \textbf{Cropping:} Starting from a raw image like Figure~\ref{fig:traffic-raw-data}, we first crop it to the shape $\perplexityinsert{imagex} \times \perplexityinsert{imagey}$,
	in order to eliminate toolbars and adapt it to the size of the graph. The background of Figure~\ref{fig:paris-stations} is obtained by the same procedure.
\item \textbf{Traffic colors extraction:}
	The colors associated with the four levels of traffic
	\[
	\texttt{colors}:=\{ \texttt{green},\, \texttt{orange},\, \texttt{red},\, \texttt{dark-red} \}
	\]
	are easily identified\footnote{The RGB value of each color is given by:
	$\texttt{green}=\perplexityinsert{green}$, $\texttt{orange}=\perplexityinsert{yellow}$, $\texttt{red}=\perplexityinsert{red}$, $\texttt{dark-red}=\perplexityinsert{darkred}$}. They seem to be used exclusively for that purpose, hence it suffices to extract the pixels having one of these colors.
 
\item \textbf{Projection on graph edges:} These pixels, once expressed in their geographical coordinates, almost perfectly overlap the metric graph from Open Street Maps. For each edge $\ee\in \E$ and each color $\cc\in \texttt{colors}$, we count the number $p_{\cc}^\ee$ of pixels of color $\cc$ that are closest to edge $\ee$. Note that the traffic color might change along an edge, in which case we give up on some local information by only considering the total traffic on the edge.

\item \textbf{Edge normalization:}
We then transform these pixel counts into proportions of traffic colors on each edge.
As the edges may remain blank at times where there is no traffic, we take as a normalizing constant the maximal amount of pixels encountered over all times~$T$ for which we collect traffic data:
\[
q^\ee_\cc(t) = \frac{p^\ee_\cc(t)}{\max_{t'\in T}\sum_{\cc'\in\texttt{colors}} p^\ee_{\cc'}(t')},\quad t\in T.
\]
In this way, $q^\ee_\cc(t)\in [0,1]$ indicates the proportion of edge $\ee$ colored with $\cc$ at time $t$, but remains null if no traffic is reported.

\item \textbf{Hourly averaging:} As the pollution information is only available every hour, we take the average of the four values of $q_\cc^\ee$ encountered every fifteen minutes, which we still denote $q_\cc^\ee$ in the sequel.

\item \textbf{Projection on graph nodes:} In our models, it is in fact simpler to localize emissions on the nodes of the graph.
For this reason, we calculate the density of each traffic color $\cc$ around a vertex $\vv\in \V$ as a weighted average on its neighboring edges $\E(\vv)$
\[
q^\vv_\cc(t)=\frac{\sum_{\ee \in \E(\vv)}a_{\ee} q^\ee_\cc(t)}{\sum_{\ee \in \E(\vv)}a_{\ee}},
\]
where $a_{\ee}$ stands for the area of the road associated to edge $\ee$, given by the product of its length $\ell_\ee$ with the number of lanes.
\end{itemize}

\subsection{Summary}

While the history of sensor and weather data can be found on archives, our script for capturing traffic images only runs in real time, since Google Traffic only provides current information. We collected all types of data on an hourly basis for a period of time comprised between December~9, 2022 and March 19, 2023.
After removing time stamps for which some data was missing, we end up with a set of acquisition times $T$, of cardinality $|T|=1712$, which we divide into a set $T_{train}$ of 1338 training times, and a set $T_\test$ of 374 testing times.

In the end, given the graph $\G=(\V,\E)$, the available information at any time $t\in T$ is of the form
\begin{equation}
\label{eq:x}
x=(\vv^\obs,z,\theta,w,(q_{\cc}^\vv)_{\cc,\vv})\in \cX=\V^m\times \R^m\times \R\times \R^2\times \R^{4|\V|}.
\end{equation}

In the next section, we present various models to estimate the pollution field from this data, using either statistical inference, linear mappings based on expert knowledge, or neural networks.

\section{Reconstruction methods}
\label{sec:concrete-methods}
We have implemented several methods of state estimation by leveraging the different information sources. Our methods give reconstructions on the metric graph $\G$, that is, we consider mappings $A:\cX\to U$ where $U$ is a space of functions defined on $\G$. Typical examples are $U = \cC(\G),\, L^2(\G)$ or $H^1(\G)$, as defined in Appendix \ref{sec:metric-graphs}. As our main interest is in assessing the effect of incorporating indirect information like the real-time traffic data, we first consider models that take only a portion of 
 $x\in \cX$ as input.

\subsection{Spatial average}
\label{sec:snapshotmean}
If we give up on all the spatially-dependent data $\vv^\obs$ and $(q_{\cc}^\ee)_{\cc,\ee}$, the reconstruction is necessarily constant over the whole domain $\G$, which yields no better choice than the average of the observed concentration values
\[
A_{\rm avg}(x)(r) =\bar z=\frac{1}{m} \sum_{i=1}^m z_i, \quad \forall r\in \G.
\]
This extremely simple reconstruction will serve as our baseline to compare more sophisticated reconstructions.
In the sequel, we will add the spatially-dependent data and view the other models as corrections to the spatial average above. This will result in spatially unbiased estimators, provided that the locations of the sensors are representative of the whole pollution field. More precisely, assuming that the stations are randomly drawn according to the uniform probability distribution $\mu$ on $\G$, the expectation over $z_1,\dots,z_m$ of the spatially-averaged error is
\[
\mathbb E_{z}\left(\int_\G \big(u(r)-A_{\rm avg}(x)(r)\big) \,d\mu(r)\right)=\int_\G u\,d\mu-\mathbb E_z\left(\frac{1}{m} \sum_{i=1}^m z_i\right)=0,
\]
and this remains true when adding to $A_{\rm avg}$ a correction of vanishing expectation.

\subsection{Best unbiased linear estimator}
\label{sec:blue}

If we only want to estimate a missing measurement $z_i$ at a given station $i\in\{1,\dots,m\}$, we may also use statistical information stemming from the history $(z_i^t)$ of the station at previous times~$t$, as well as the observations from other stations $j\neq i$ in the present and the past, denoted respectively $z_j$ and $(z_j^t)$. For $T_\train$ the set of training times, define the empirical average
\[
\<z_i\>:=\frac{1}{|T_\train|}\sum_{t\in T_\train}z_i^t
\]
and empirical covariance matrix $K\in\R^{m\times m}$ with entries
\[
K_{i,j}:=\Big\<\big(z_i-\<z_i\>\big)\big(z_j-\<z_j\>\big)\Big\>=\<z_iz_j\>-\<z_i\>\<z_j\>.
\]
Any unbiased linear estimator $\tilde z_i$ of $z_i$ is of the form
\[
\tilde z_i=\<z_i\>+\sum_{j\neq i}c_j\big(z_j-\<z_j\>\big),
\]
for some coefficients $(c_j)_{j \neq i}$. Let $c\in \R^m$ be the vector with coordinates $c_j$ for $j\neq i$ and $c_i=-1$.
Then the best linear unbiased estimator (BLUE) is obtained by optimizing the averaged squared error
\[
\argmin_{(c_j)_{j\neq i}}\Big\<\big(\tilde z_i- z_i\big)^2\Big\>
=\argmin_{(c_j)_{j\neq i}} \,c^\top Kc
=\big[(K_{j,k})_{j,k\neq i}\big]^{-1}(K_{j,i})_{j\neq i},
\]
where $(K_{j,k})_{j,k\neq i}$ and $(K_{j,i})_{j\neq i}$ are seen as a matrix in $\R^{(m-1)\times (m-1)}$ and a vector in $\R^{m-1}$.

If the set of training times $T_\train$ is large enough, we expect an ergodicity property of the form $\<z_i\>\approx \E(z_i)$ to hold. For this reason, BLUE should be a near minimizer of the expected squared error, given the available data. Therefore, in the numerical experiments, we will evaluate the different methods $A$ by comparing $A(x\setminus \{z_i\})(z_i)$ and $z_i$, and the error $|\tilde z_i-z_i|^2$ will act as an optimality benchmark.

It should be emphasized that BLUE itself is not a valid reconstruction method, since it requires statistical information which is accessible only at the locations of the stations $\vv_i^\obs$. Hence this estimator cannot be computed at any point $r\in \G$ of the graph domain.

\subsection{Kriging}
\label{sec:kriging}

In order to transform BLUE into a reconstruction method, one needs to propose a surrogate for the correlation between any two points in the graph. Moreover, as we don't know the average pollution at all points of the graph, we proceed without subtracting spatial averages $\<z_i\>$ in this subsection, in contrast to the previous one. Therefore, we consider the Gram matrix of normalized second-order moments
\[
G_{i,j}=\frac{\<z_iz_j\>}{\sqrt{\<z_i^2\>\<z_j^2\>}}.
\]

\begin{figure}[ht]
          \centering
          \includegraphics[width=0.6\textwidth]{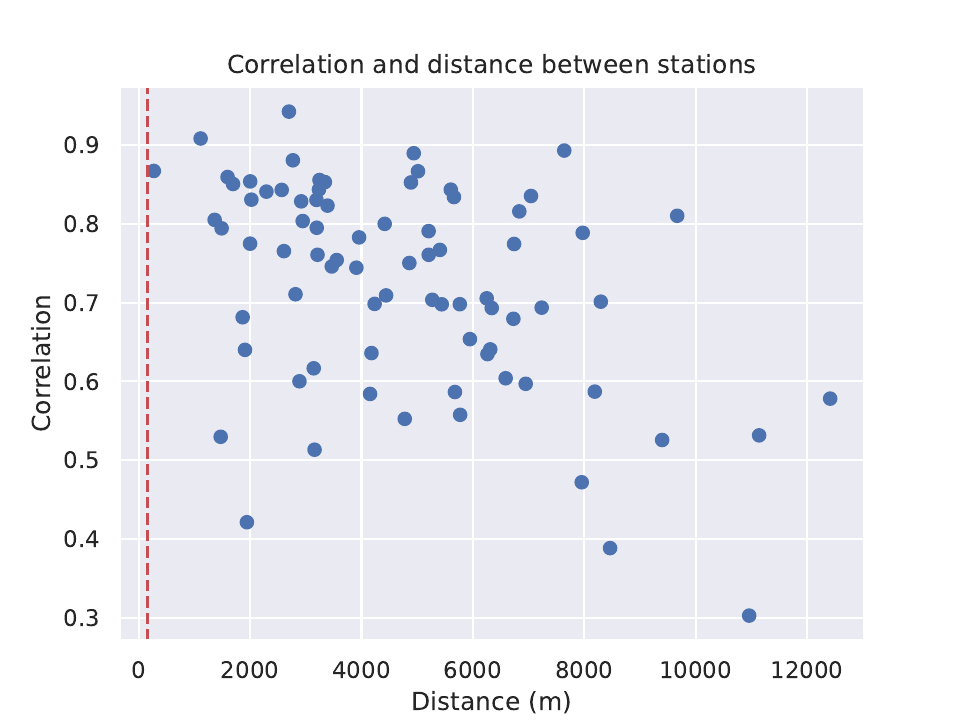}
          \caption{Correlation between stations as a function of the distance. The vertical slashed red line marks the maximal separation between vertex and station (165m) which still lays in the zone of high correlation.}
          \label{fig:correlation}
\end{figure}

Taking the positions $\vv^\obs$ of the stations into account, we observe that each entry $G_{i,j}$ partly depends on the distance $|\vv_i^\obs-\vv_j^\obs|$ between the stations, see Figure~\ref{fig:correlation}. A typical choice of approximant is the Gaussian kernel
\[
G_{i,j}\approx \hat G_{i,j}:=C\exp\left(-\frac{|\vv_i^\obs-\vv_j^\obs|^2}{2\sigma^2}\right)+(1-C)\delta_{i,j},
\]
with parameter values $C=0.968$ and $\sigma=33.4{\rm km}$ obtained by fitting the station data in our case.

\begin{remark}
The fact that $C<1$ can be interpreted as the presence of random noise $\eta_i$ on the measurements $z_i=u(\vv_i^\obs)+\eta_i$. As a safety check, one may notice that the average relative error
\[
\frac{\<\eta_i^2\>}{\<u(\vv_i^\obs)^2\>}=\frac{\<z_i^2\>}{\<u(\vv_i^\obs)^2\>}-1\approx \frac{1}{C}-1=3.31\%
\]
is effectively much smaller than the uniform error guarantee $\|\eta_i/|u(r_i^\obs)\|_{L^\infty}\leq 15\%$ that we discussed in Section \ref{sec:pollution-sensors}. Adding the matrix $(1-C)I$ ensures that $\hat G$ has ones on its diagonal, as expected of a correlation matrix, and regularizes the system, by making the inversion of $\hat G$ stable. More practically, the reconstructed value $A(x)(\vv_i^\obs)$ will not be exactly $z_i$, but rather an average of the measurements close to $\vv_i^\obs$.
\end{remark}

Let $r\in \G$, we again examine a linear model
\[
A_{\rm krig}(x)(r)=\sum_{i=1}^m c_i^r z_i,
\]
with coefficients $c^r\in \R^m$ to be determined.
This estimator is unbiased if and only if $\sum_{i=1}^m c_i=1$, and in that case we can write it as a correction to the temporal or spatial average
\[
A_{\rm krig}(x)(r)=\<A_{\rm krig}(x)(r)\>+\sum_{i=1}^m c_i^r (z_i-\<z_i\>)=\bar z +\sum_{i=1}^m \left(c_i^r-\frac{1}{m}\right)(z_i-\bar z).
\]
By analogy with BLUE, we thus define the weights as the renormalized solution of a system of correlation equation
\[
c^r=\frac{\hat c^r}{\sum_{i=1}^m \hat c_i^r},\qquad \hat c^r =\hat G^{-1}g^r, \qquad g_i^r=C\exp\left(-\frac{|\vv_i^\obs-r|^2}{2\sigma^2}\right).
\]
Although there are no optimality guarantees, we expect kriging to have intermediate performance when compared to the spatial average baseline, and to the ideal BLUE reconstruction.
However, due to the important spacing between peripheral stations, we only observe an improvement in the central part of Paris. The insufficient density of pollution measurements calls for models involving other sources of information, such as traffic data. This is the objective of the next two subsections.

\subsection{Source model}
\label{sec:source-model}

The simplest way to incorporate traffic data consists in using only local values $q_\cc^\ee$ for estimating the pollution on an edge $\ee\in \E$, or $q_\cc^\vv$ for a node $\vv\in \V$. As we projected the station locations on~$\V$, we focus on the latter case here. We call such a method a source model, since it directly maps the sources of emission to pollution values.

We opt for a linear model acting as a correction on the spatial average baseline:
\begin{equation}
A_{\rm src}(x)(\vv)=\bar z+\sum_{\cc \in\texttt{colors}}\alpha_\cc (q_\cc^\vv-\bar q),
\label{source_linear}
\end{equation}
where we substracted the spatial average of traffic $\bar q$ for unbiasedness.
The vector of coefficients $\alpha\in \R^4$ is found by solving a LASSO problem
\[
\min_{\alpha\in \R^4} \sum_{t\in T_\train}\sum_{i=1}^m |z_i^t - A_{\rm src}(x(t))(\vv^\obs_i)|^2 + \lambda \Vert \alpha\Vert_1,
\]
and we perform a cross-validation to estimate the optimal parameter $\lambda$, in order to prevent overfitting.

Alternatively, we can also write nonlinear variants, of the form
\begin{equation}
A_{\rm src}(x)(\vv)=\bar z+ \cT_\alpha\left((q^\vv_\cc),\theta, w\right),
\label{source_nonlinear}
\end{equation}
which may take into account other sources of information like temperature $\theta$ and wind $w$.
Here, $\cT:\R^{\#\alpha} \times \R^4 \times \R \times \R^2 \rightarrow \R$ can be a \textit{polynomial} combination of the inputs $\left((q^\vv_\cc),\theta, w\right)$, a \textit{neural network}, or any other nonlinear mapping.
The set of parameters $\alpha$ is no longer associated to the four traffic colors, but still needs to be learned via a LASSO regression.

All such models rely on the assumption that pollution depends on its sources in a very localized manner. However, the traffic charts $(q_\cc^\vv)_{\vv\in \V}$ exhibit sharp  variations from one node to its neighbors, which incites to smooth the emissions before applying the above methods. This is the purpose of the following section, which attempts to model such a diffusive behavior.

\subsection{Physical modeling}
\label{sec:physical}

An important inspiration for reconstruction methods comes from the physical modeling of pollution dispersion. In our setting, we resort to building a \emph{quantum graph}, that is, we endow our metric graph $\G$ with a differential operator acting on functions from functional spaces such as $L^2(\G)$ or $H^1(\G)$, as defined in Appendix \ref{sec:metric-graphs} (we also refer to \cite{BK2013} for more details and references). Our approach can be summarized as follows:

\paragraph{Elliptic equation:} One can first model pollution with a time-independent elliptic equation, by assuming that all time-dependent parameters have sufficiently slow variations, here over the course of an hour, for the pollution field to reach a steady state. For any point $r\in \G$, the pollutant concentration is modelled by a function $u: \G \to \bR$ solution to the diffusion-reaction equation
\begin{equation}
\label{eq:strong-form-elliptic}
\mathcal P(u):=- \frac{\dd}{\dr} \cdot \left(a(r) \frac{\dd}{\dr} u(t, r)\right)+h(r)u(r) = q(r),\qquad r\in  \G
\end{equation}
which we choose to complement with ``Newmann-Kirchoff'' conditions on the vertices, that is,
\begin{equation}
\label{eq:newmann-kirchoff}
\sum_{\ee\in \E(\vv)}\frac{\dd u}{\dr}\Big\vert_\ee (\vv) = 0,\qquad  \vv \in \V,
\end{equation}
expressing the conservation of the quantity of pollutant at every crossroad $\vv \in \V$.
Here, $\E(\vv)$ denotes the edges having $\vv$ as an endpoint, and the derivatives are assumed to be taken in the directions away from the vertex.

In equation~\eqref{eq:strong-form-elliptic}, the function $a\in L^\infty(\G)$ is an effective diffusion coefficient, which takes into account turbulent dissipative effects. The absorption coefficient $h\in L^\infty(\G)$ models the leakage of pollutants from the streets to the higher atmosphere, as well as chemical reaction, in particular between ${\rm NO}_2$ and other nitrogen oxides, which are not measured by the sensors. Lastly, the source term $q\in L^2(\G)$ models all possible emissions of pollutant, which in the case of Paris essentially come from traffic, local heating, and industrial and urban activity outside the city. As we only have access to local traffic data, we assume that the other sources are spatially constant, and average them out by solving $\mathcal P(u-\bar u)=q-\bar q$, where $\bar u$ is the spatial average of the pollutant concentration, estimated by $A_{\rm avg}(x)=\bar z$, and $q-\bar q$ corresponds to the variations of traffic around its spatial average, computed through the procedure from Sections~\ref{sec:traffic-preprocess} and \ref{sec:source-model}.

\begin{remark}
Equation~\eqref{eq:newmann-kirchoff} has a similar effect as Newmann conditions at the borders of the spatial region under consideration, here the rectangle contour of Figure~\ref{fig:paris-stations}. Therefore the only exchanges with the exterior of this region are contained in the source term $q$. As this no-flux condition only give a very rough approximation of the solution close to the border, in the numerical experiments of Section~\ref{sec:numerical-tests}, we will concentrate on the accurate prediction of the pollution in the central part of the city.
\end{remark}

\paragraph{Variational formulation:} The operator $\mathcal P(u)$ in \eqref{eq:strong-form-elliptic} is defined for functions $u\in H^2(\G)$,
but the equation can be stated in a weak form, which only requires that $u \in H^1(\G)$. 
Multiplying \eqref{eq:strong-form-elliptic} by a sufficiently smooth test function $v\in H^1(\G)$, and using the Kirchoff-Neumann boundary conditions, it follows that the corresponding weak formulation of the problem is to find $u\in H^1(\G)$ such that
\begin{equation}
\label{eq:weak-form-elliptic}
\mathfrak{b}(u, v) = \mathfrak{f}(v), \qquad  v \in H^1(\G)
\end{equation}
where $\mathfrak{b}$ is the symmetric bilinear form defined as
\[
\begin{array}{rccl}
 &H^1(\G)^2 &\to &\bR \\
\mathfrak{b}:&(u, v) &\mapsto &
\sum_{\ee \in \E}
\displaystyle{\left\lbrace
\int_\ee a(r)\frac{\dd u}{\dr}(r) \frac{\dd v}{\dr}(r) \dr + \int_\ee h(r) u(r)v(r)\dr
\right\rbrace}
\end{array}
\]
and $\mathfrak{f}: v\in H^1(\G)\mapsto \sum_{\ee\in \E} \int_{\ee} q(r) v(r)\dr$
is a continuous linear form.

Assuming that $a(r)\geq a_0 >0$ and $h(r)\geq h_0 > 0$ for $r\in \G$ a.e., we see that $\mathfrak{b}$ is continuous and coercive in $H^1(\G)$ with coercivity constant $\min(a_0, h_0)$, and continuity constant $\max(\Vert a \Vert_{L^\infty(\G)}, \Vert h \Vert_{L^\infty(\G)})$. By the Lax-Milgram theorem, problem \eqref{eq:weak-form-elliptic} admits a unique solution $u\in H^1(\G)$.

\paragraph{Discretization:} In our numerical tests, we discretize the equation with $\bP_1$ finite elements, that is, continuous functions whose restriction to any edge is affine. We describe below the main guidelines, and refer to \cite{AB2018} for further details and a complete analysis.

We define the set of hat functions $\{\varphi_\vv\}_{\vv\in \V}$ by $\varphi_\vv( \vv') = \delta_{\vv, \vv'}$ for any vertices $\vv,\vv'\in \V$, and
\[
\forall x_\ee \in [0, \ell_\ee],\quad
\varphi_\vv(x_\ee)=
\begin{cases}
1-\frac{x_\ee}{\ell_\ee}, \quad &\text{ if } \ee \in \E(\vv),\\
0 , \quad &\text{ if } \ee \not\in \E(\vv),
\end{cases}
\]
for any edge $\ee \in \E$.
Fixing our finite element space $\bP_1 = \vspan\{\varphi_\vv\}_{\vv\in \V} \subset H^1(\G)$, we search for the Galerkin solution $\hat u = \sum_{\vv\in \V} c_\vv \varphi_\vv \in \bP_1$ such that
\[
\mathfrak{b}(\hat u, \hat v) = \mathfrak{f} (\hat v), \qquad  \hat v\in \bP_1.
\]
Gathering the expansion coefficients of the solution in the vector $\textbf{c} = \{c_\vv\}_{\vv\in \V}$, we obtain the linear system of equations
\begin{equation}
\bB\,\textbf{c} = \textbf{f}
\label{eq:galerkin_system}
\end{equation}
with $\bB = (\mathfrak{b}(\varphi_\vv, \varphi_{\vv'}))_{\vv,\vv'\in \V}$ and $\textbf{f} = (\mathfrak{f}(\varphi_\vv))_{\vv\in \V}$. Again by Lax-Milgram theory, this system is invertible, which allows to compute the solution $\hat u$.

\paragraph{Reduced models:} Unfortunately, solving equation \eqref{eq:galerkin_system} is expensive, given the size $|\V|\approx 10^4$ of the graph, so we cannot afford to find $\hat u$ at each time step. In order to mitigate the computational cost, we rely on model order reduction techniques, which have received much attention in the context of parametrized elliptic PDEs \cite{CD2015acta,CDS2011, EPR2010, HRS2015, RHP2007, ZKA2019}. Here, the parameters would be the diffusion $a$, the reaction $h$, and the right-hand side $q$.
We consider three reconstructions methods.
\begin{enumerate}
\item \textbf{Eigenstates of the graph Laplacian:} One option consists in taking as a reduced model the subspace $V_n\subset H^1(\G)$ spanned by the $n$ first eigenfunctions of the Laplacian in $\bP_1$. As this operator is self-adjoint and coercive, it admits a spectral decomposition with positive eigenvalues, and the coefficients of the eigenstates in the basis $\{\varphi_\vv\}_{\vv\in \V}$ are the eigenvectors of $\bB$.
Assuming that the diffusion and reaction coefficients $a$ and $h$ are constants calibrated in a pre-processing phase, we define the reconstruction mapping $A: \cX \to H^1(\G)$ by taking $\hat u=A(q)$ the Galerkin projection of $u$ onto $V_n$, that is, by searching $\hat u\in V_n$ solution to
\[
\mathfrak{b}(\hat u, \hat v) = \mathfrak{f} (\hat v), \qquad \hat v\in V_n,
\]
which is simply a diagonal system in the eigenstate basis. We then plug $\hat u$ instead of $q$ in equation \eqref{source_linear} or \eqref{source_nonlinear}, and learn the coefficients associated to each color, or the more general parameters $\alpha$.

\item \textbf{Principal components of traffic data:}
Starting with the whole history of traffic data $(q(t))_{t\in T}\in \R^{|T|\times |\V|}$, we can also perform a singular value decomposition to find the $n$ first modes $q_1,\dots,q_n$, compute the solutions to $\mathcal P(u_k)=q_k$, and assemble them in a reduced space $V_n=\vspan\{u_1,\dots,u_n\}$. In this way, we expect $V_n$ to better capture physical properties of the pollution field, such as strong correlations along a large avenue.

As full-order solves remain costly, we resort to a convolution with a gaussian kernel:
\[
(u_k)_\cc^\vv=\frac{\sum_{\vv'\in \V}e^{-d_{\vv\vv'}^2/2\delta^2}(q_k)_\cc^{\vv'}\phi_{\vv'}}{\sum_{\vv'\in \V}e^{-d_{\vv\vv'}^2/2\delta^2}}
\]
where $d_{\vv\vv'}=|\vv-\vv'|$ is the distance in $\R^2$ (which is equivalent, up to constants, to the distance on the graph). We set $\delta=400\, \text{m}$, after observing that pollution data is optimally correlated to regularized traffic information for $\delta$ close to this value.

In our experiments, we perform the smoothed projections $q\mapsto \hat u=\sum_{k=1}^n \<q,q_k\>u_k$ into a different reduced space for each of the four traffic colors.
After this operation, we can apply any of the strategies described in \cref{sec:source-model} to $\hat u$ instead of $q$.
\end{enumerate}

These two methods regularize the traffic data, but they do not exploit the information from the pollution sensors, apart from the average value $\bar z$.
In order to assimilate data of both types, it is possible to use a combined least-squares fit of the form
\[
A(x)=\argmin_{\hat v\in V_n}\Vert z - \hat v(\vv^\obs)\Vert^2_2+\lambda'\|q-\hat q\|_{\ell^2(\V)}^2,
\]
where $\lambda'>0$ balances the contributions of $z$ and $q$, and $\hat q=\mathcal P(\hat v)=\sum_{k=1}^n \hat c_kq_k$ for the coefficients $\hat c\in \R^n$ such that $\hat v=\sum_{k=1}^n \hat c_ku_k$. However this approach did not perform well in practice, so we did not include it in the numerical experiments.

In the last method, if $m\geq n$ and $\lambda$ tends to $0$, the prediction $\hat u$ does a least squares fit of $u$ at the available measurement points $\vv_i^\obs$. In general, it is possible to enforce $\hat u(\vv^\obs)=u(\vv^\obs)$ by applying a correction to the prediction. This post-process, called \textit{Parameterized Background Data-Weak} method, was originally introduced in \cite{MPPY2015} and has been analyzed and extended in a series of papers such as \cite{BCDDPW2017, CDDFMN2020, CDMN2022, CDMS2022}.
The whole approach has found numerous applications, including pollution dispersion \cite{HCBM2019}.

It would of course be possible to gain in accuracy, by considering more refined equations for pollution dispersion, which capture additional physical properties, and thus by encoding these properties into the reduced space $V_n$. One could for instance think of advection by wind, vertical fluxes or stratification of the atmosphere depending on the temperatures, changes in the chemical equilibrium between ${\rm NO}$ and ${\rm NO}_2$ caused by cloud coverage and precipitations~\cite{li2020quantifying}, as well as local turbulent effects near the sensor stations.
Note that, if nonlinear equations are involved, $V_n$ can be a nonlinear approximation space defined through a chart of $n$ parameters, and approximation guarantees are more difficult to obtain \cite{CDMS2022}.

\subsection{Super-Learning as a collaborative approach}
\label{sec:super-learning}

To gain in accuracy over each individual model, one can combine a set of $p$ available mappings $A_1, \dots, A_p$ coming from the previous methods, and build a super-learner
\[
\begin{array}{rccc}
&\cF(\cX, U)^p &\longrightarrow &\cF(\cX, U) \\
\cS:& ( A_1, \dots, A_p) &\longmapsto &\cS(A_1, \dots, A_p),
\end{array}
\]
where $\cF(\cX, U)$ denotes the set of functions from $\cX$ to $U$. The most simple merger, usually called aggregator in statistics, amounts to taking a linear combination
\[
\cS_\omega(A_1, \dots, A_p) = \sum_{i=1}^p \omega_i A_i,
\]
for some weights $\omega=(\omega_1, \dots, \omega_p)$ expressing the confidence in each individual model.

More sophisticated strategies involve nonlinear combinations and compositional structure. One could think of using a first model to obtain a rough estimation, and compose it with a second model performing refinements based on its output. This is already an underlying idea in our constructions, where we start with the spatial average, and add spatially-dependent corrections. The physical models involve one more compositional step, since they are of the form $A_{\rm src}\circ \hat u(q)$.

Neural networks constitute another prominent example of nonlinear super-learners: one could treat $A_1(x),\dots,A_p(x)$ as inputs, and train the parameters $\omega$ by minimizing an empirical loss. We would like to emphasize here that properly training the super-learner requires to implement a nested leave-one-out strategy: one should first train each parametrized submodel by leave-one-out, and then optimize the neural network with another leave-one-out step, in order to avoid overfitting. As a consequence, at least two observation points are removed from the training set of the submodels, which may cause a loss of accuracy, especially when the number $m$ of observations is small.

In our application, the neural network super-learner performed slightly worse than its linear counterpart, which should come as no surprise in view of the above observation.

\section{Reconstruction benchmarks and Leave-One-Out}
\label{sec:rigorous}

There are several ways to quantify the quality of a reconstruction map $A:\cX \to U$. Ideally, given a state $u\in U$ and the associated  observations $x\in \cX$, one would like to find $A$ such that the error $\Vert u - A(x)\Vert_U$
is as small as possible. Assuming that $(u,x)$ is a random variable with distribution $\pi\in {\rm Prob}(U\times \cX)$,
we define the performance of $A$ as the $L^2$ norm of the error
\[
e(A)^2:=\int_{U\times \cX}\Vert u - A(x)\Vert_U^2 \,d\pi(u,x),
\]
which acts as a good compromise between the worst-case error and the average error. In addition, although the state $u$ has in principle $H^2$ regularity, we asses the spatial error also in $L^2$, that is, we take $U=L^2(\G)$.
Unfortunately, finding $A$ minimizing $e(A)$ is out of reach for several reasons.
First, we don't know the distribution $\pi$, nor even its support, which is the set of all possible states and observations. Second, given $u\in U$, we cannot evaluate 
$u(r)$ at any point $r\in \G$, making the computation of $\Vert u - A(x_u)\Vert_U$ intractable. 

Concerning the first issue, as we have access to hourly data on a large period of time, we can replace the integral over $\pi$ by an empirical average
\[
e(A)^2\approx \frac{1}{T_\test}\sum_{t\in T_\test} \|u(t)-A(x)(t)\|_U^2
\]
over the set $T_\test$ of $374$ test times. Assuming that these states are independent, this approximation induces an error of order
\[
\frac{\mathbb E(\|u\|_U^2)^{1/2}}{\sqrt{|T_\test|}}\approx \frac{41}{19.3}\approx 2.1 \,{\rm \mu g/m^3},
\]
which is totally acceptable in view of the noise level on the measurements.

The second obstacle is more tricky, because we only know $u$ at a very limited number $m$ of fixed positions, and because these observations are also needed for constructing $A$. Ignoring the last issue leads to an systematic underestimation of $e(A)$, as we detail below.

Assume that the observation points $\vv_i^\obs$ are distributed uniformly at random on $\G$, define the discrete semi-norm
\[
\|u\|_{m}^2=\frac{1}{m}\sum_{i=1}^m |u(\vv_i^\obs)|^2
\]
corresponding to an empirical version of $\|u\|_U^2$, and consider map $A$ solution to
\[
\min_{A:\cX\to V_n\text{ linear}} \frac{1}{T_\train}\sum_{t\in T_\train} \|u-A(x)\|_m^2
\]
for some linear space $V_n\subset U$ of dimension $n$. This setting is valid for most of our methods, with $V_n$ the set of constant functions in the case of $A_{\rm avg}$ (of dimension $n=1$), but also $V_n=\vspan\{r\mapsto c_i^r\}_{1\leq i \leq m}$ in the case of $A_{\rm krig}$ (of dimension $n=m$), and $V_n$ the reduced basis in the methods based on physical modeling.

Then, for $A^*$ the optimal map taking values in $V_n$
\[
A^*=\argmin_{A':\cX\to V_n\text{ linear}} \mathbb E(\|u-A'(x)\|_U^2)=\mathbb E_\pi(u|x),
\]
we obtain, by applying Pythagoras theorem both for $\|\cdot\|_U$ and $\|\cdot\|_m$,
\[
e(A)^2=\mathbb E(\|u-A(x)\|_U^2)\geq \mathbb E(\|u-A^*(x)\|_U^2)
= \mathbb E(\|u-A^*(x)\|_m^2)
\geq \mathbb E(\|u-A(x)\|_m^2),
\]
where the central equality comes from the assumption that the $\vv_i^\obs$ are random. This proves that $\|u-A(x)\|_m^2$ is a biased estimator for $e(A)^2$ as soon as one of the inequalities is strict, that is, as soon as $A(x)\neq A^*(x)$.
Note that separating the training and test data by splitting the set of time indices $T$ is not sufficient, since the algorithm will then fit its prediction to the station locations, without generalization guarantees to the rest of the domain $\G$.

As a consequence, we must separate the stations into a training set and test points. In order to compute an unbiased estimator of $e(A)^2$ with minimal variance, while keeping the maximal number of stations in the training set, we proceed to leave-one-out cross-validation.
This procedure is very standard and has been used in other works on pollution reconstruction, such as~\cite{criado_data_2023}.
For $1\leq i \leq m$, denote
\[
\|u\|_{m\setminus i}^2=\frac{1}{m-1}\sum_{j\neq i} |u(\vv_j^\obs)|^2\quad\text{and}\quad A_i=\argmin_{A:\cX\to V_n\text{ linear}} \frac{1}{T_\train}\sum_{t\in T_\train} \|u-A(x)\|_{m\setminus i}^2.
\]
Assuming that the station locations are independent random variables, the cross-validation estimator of the error
\[
e_{\rm CV}(A)^2:=\frac{1}{T_\test}\sum_{t\in T_\test} \frac{1}{m}\sum_{i=1}^m |u(\vv_i^\obs)-A_i(\vv_i^\obs)|^2
\]
is unbiased, since
\[
\mathbb E(e_{\rm CV}(A)^2)=\mathbb E(|u(\vv_i^\obs)-A_i(\vv_i^\obs)|^2)=\mathbb E(\|u-A_i(x)\|_U^2)=e(A),
\]
with the difference that $x$ contains only $m-1$ direct evaluations of $u$ this time.

\section{Numerical results}
\label{sec:numerical-tests}

We have implemented and tested numerous variants and combinations of the models from \cref{sec:concrete-methods}. This was done thanks to a Python code we have developed, which can be found at \url{https://github.com/agussomacal/CityPollutionModeling}. The interested user could add its own models for further testing. In this section, we summarize the most important results that emerge from our tests. We report on the performance of  the following reconstruction strategies:
\begin{itemize}
\item \textbf{Spatial average}: We take a simple spatial average, as in \cref{sec:snapshotmean}. The resulting error serves as a baseline, which we expect to beat with the other more sophisticated models.

\item \textbf{BLUE}: As explained in \cref{sec:blue}, BLUE can be seen as an estimate of the optimal linear reconstruction method. It can be used as a benchmark of the best performance that we can expect of linear methods. Note that, in principle, nonlinear strategies could be more accurate than BLUE. However, we will see in our experiments that none of our methods achieves such accuracy.

\item \textbf{Kriging}: We apply the kriging method depicted in \cref{sec:kriging} with an exponential kernel. The parameters $\sigma$ and $C$ are obtained by fitting an exponential to the correlation between training stations (that is, we do not into account correlations with the station that is set aside for testing) as a function of their distance (see \cref{fig:correlation}).

\item \textbf{Source}: We apply a linear {source model}, as described in \cref{sec:source-model}, with temperature $\theta$ and wind $w$ as extra regressor variables.

\item \textbf{Physical-PCA}: We apply the second {physical model} from \cref{sec:physical}, using a gaussian kernel to smooth the node traffic data. The four reduced spaces $V_n$, associated to the four traffic colors, each consist of the first $10$ principal components of the corresponding traffic data, as observed in the training set. After the smoothing and projection operations, we assemble the variables $\theta$, $w$ and the $q_\cc^\vv$ on each node $\vv$ into a vector $s=(q_\green^\vv,q_\yellow^\vv,q_\red^\vv,q_\dark^\vv, \theta, w)\in \R^6$ and build a \textit{polynomial} model of degree $2$:
\[
\cT_\alpha(s) := \sum_{j=1}^6 \alpha_{j}s_j + \sum_{j,k=1}^6 \alpha_{jk}s_js_k,
\]
where $\alpha\in \R^{42}$ is computed following the lines of \cref{sec:source-model}.

\item \textbf{Physical-Laplacian}: We apply the first {physical model} from \cref{sec:physical}, with the reduced space consisting of the first $5$ eigenvectors of the graph laplacian. After projection into the subspace, we take the $4$-colour traffic values on each node $(q_\cc^n)_{\cdot,i}$ and obtain their \textit{degree-$3$ polynomial} combinations. Finally we apply a \textit{neural network} consisting of two hidden layers of $20$ neurons each and a ReLU activation function. The neural network is trained with ADAM optimizer with early stopping to prevent overfitting.

\item \textbf{Ensemble}: We apply an {ensemble} model, as described in \cref{sec:super-learning}, combining the {Kriging} method $A_{\rm krig}$, the {Source} model $A_{\rm src}$ and the {Physical-Laplacian} model $A_{\rm lapl}$. We train each of them separately and compute the following linear combination:
\[
A_{\rm ens}(x)(r)=\omega(r) A_{\rm krig}(x)(r)+\frac{1-\omega(r)}{2}A_{\rm src}(x)(r)+\frac{1-\omega(r)}{2}A_{\rm lapl}(x)(r),
\]
with a weight function $\omega(r)=\exp(\min_{1\leq i\leq m}|r-\vv_i^\obs|/\delta)$, where $\delta=800\,\text{m}$. Essentially, we favour Kriging when $r$ is close to one of the sensor stations, and average the predictions of models using local or global traffic information otherwise.
\end{itemize}

\begin{figure}[ht]
    \centering
    \includegraphics[width=0.95\linewidth]{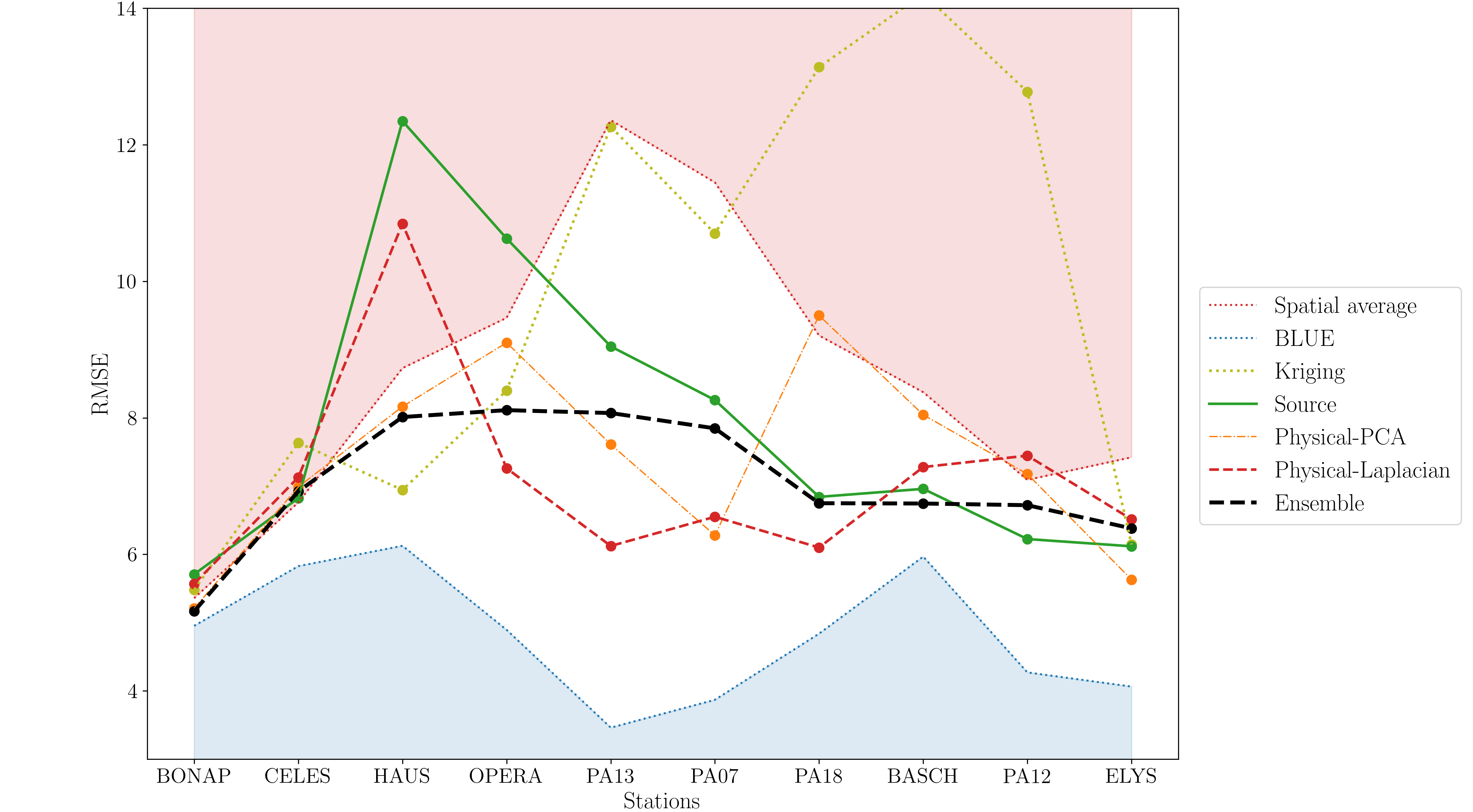}
    \caption{Root mean square error on tested stations for the different proposed methods}
    \label{fig:errorplots}
\end{figure}

In \cref{fig:errorplots}, we show the root mean square error (in $\mu \text{g/m}^3$) for each model's predictions on the test times $T_\test$ and tested stations $i$:
\[
e_{\rm RMSE}(A, i):=\left(\frac{1}{T_\test}\sum_{t\in T_\test} |u(\vv_i^\obs)-A_i(\vv_i^\obs)|^2\right)^{1/2}.
\]
Note that the cross-validation error $e_{\rm CV}(A)$ from \cref{sec:rigorous} is just the $\ell^2$-average of these errors over all stations.
 For the tests, we only keep the $10$ stations located in the interior of Paris. The remaining $3$ are set aside because they have a significant proportion of missing values (in average $10\%$ of the data is lacking in each of these stations), and because they lay close to the border of the image, making the surrounding traffic information incomplete.

The shaded blue area is the region corresponding to errors smaller than the reference \textit{BLUE} model. On the opposite side, the shaded red area marks situations in which the prediction is worse that the \textit{spatial average} baseline. The white margin in between indicates the region where we expect feasible improvements.

We first notice that using a linear \textit{source model} already yields reliable improvements with respect to the \textit{spatial average} baseline. It fails, however, in HAUS and OPERA stations due to the absence of traffic information around the former, as the Google Maps symbol for the Paris Opera is drawn over the location of the sensor. This affects the predictions on both stations but most prominently on HAUS. This problem can be alleviated if we average traffic information over a bigger region, as done in \textit{Pysical-PCA} thanks to the Gaussian smoothing, at the expense of losing precision on other stations like PA18. 

The \textit{Kriging} model manages to produce enhanced predictions in both HAUS and OPERA stations because of their proximity and correspondingly high correlation in pollution values. However, for other stations, especially those further from the center, the performance highly deteriorates.

\begin{figure}[ht]
    \centering
    \includegraphics[width=0.85\linewidth]{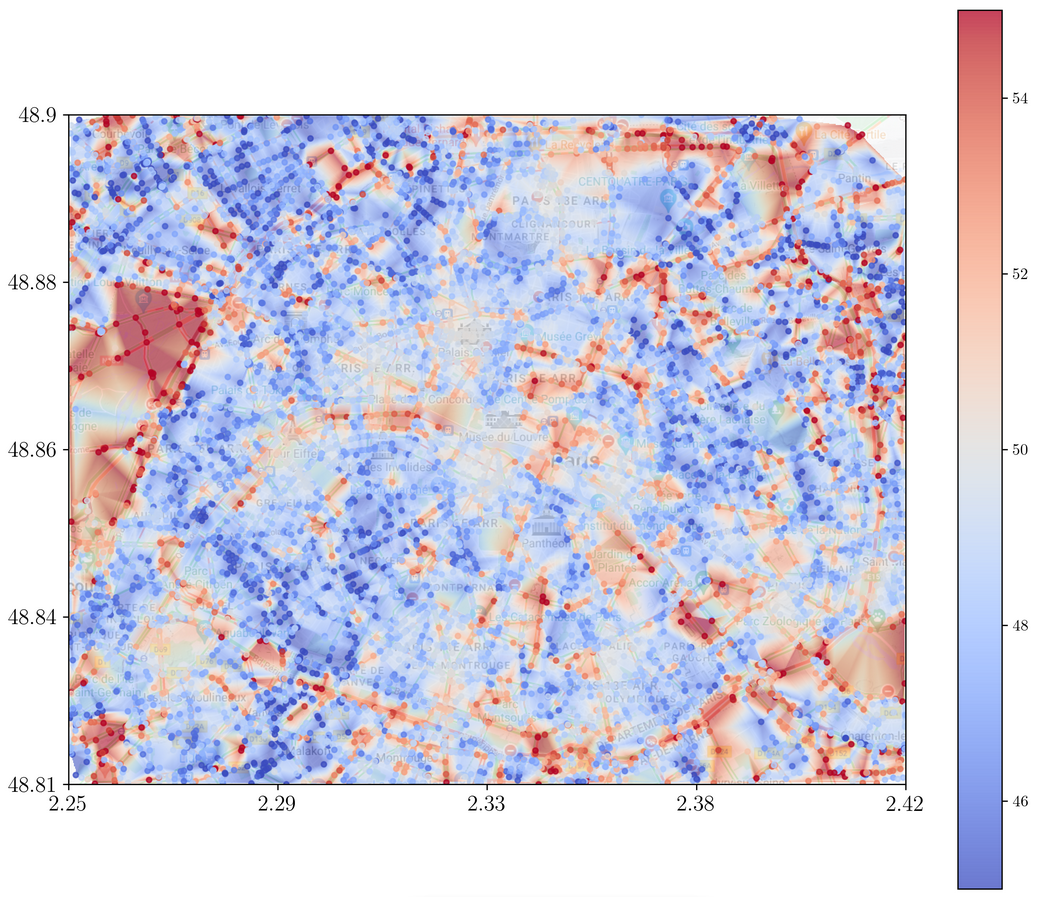}
    \caption{Pollution map for the \textit{ensemble} model at $8$am on March 1st, 2023. Based on the predictions on the node, one can linearly extrapolate pollution values even outside of the graph edges. Note that the fine variations in pollutant concentration (between 45 and 55 $\mu \text{g/m}^3$) seem to trace the main circulation axes.}
    \label{fig:pollution-map}
\end{figure}

With the \textit{Physical-Laplacian} model, we get further improvements in $6$ stations compared to the linear \textit{source model}, while losing some advantage in the remaining $4$. Finally, the \textit{ensemble} method, by combining two traffic models and the \textit{Kriging} method, manages to exploit the advantages of each in a pretty decent way. It yields predictions that beat or equal the \textit{spatial average} baseline on all stations, and that reach the best average error among all our tested methods. In \cref{fig:pollution-map}, we show an example of pollution maps generated with this last model.

\section{Conclusion and future works}

In this work, we have showed that it is possible to leverage pollution sensor data, meteorological information and Google Traffic images to create pollution maps in real time. In particular, we explained how to build statistical, physics-based and ensemble reconstruction strategies by posing the problem of pollution state estimation on metric and quantum graphs. Furthermore, the right combination of these techniques produced systematic improvements over the proposed baselines, namely the \textit{Spatial average} and \textit{Kriging}.

Neither our linear reconstruction strategies nor our nonlinear ones could beat the BLUE benchmark that indicates the accuracy of the best linear estimator. We conjecture that this is due to the limited amount of stations giving us spatial information on the pollution field, and to the indirect nature of traffic data. Regarding the last point, even though the volume of traffic information is large, it still remains of reduced utility. This is due to the fact that we only measure the fluidity of traffic, instead of the actual amount of passing vehicles, which is the relevant variable directly impacting emissions. One can then hope to obtain further improvements by following a similar approach with better suited data.

Another limitation is the unavailability of local pollution averages for the city of Paris. Having access to this kind of data through intensive measurement campaigns lasting a few weeks, but employing hundreds to thousands of sensors, as done in \cite{criado_data_2023}  for the city of Barcelona, would give a much more precise baseline, and allow to learn corrections to the local average instead of the global one. 

Finally, in setting the problem on the graph, we did not take into account the vicinity of open spaces like parks or rivers, nor the topology and variations in altitude. This is particularly visible in \cref{fig:pollution-map}, where the parks of Boulogne and Vincennes are colored in red because of the surrounding highways, and the absence of small internal streets.
We leave the inclusion of such relevant features to future studies. 

\bigskip

\textbf{Acknowledgments and disclosure of funding:}
The authors would like to thank Prof.~Albert Cohen, Prof.~Joubine Aghili, Dr.~Rachida Chakir, Dr. Vivien Mallet, and Dr.~Fabien Brocheton for fruitful discussions, and for preliminary work on the extraction of applicable data.

This work was done in the framework of the research project ``Models and Measures'' funded by the Parisian City Council (Emergences grant program). In addition, Matthieu Dolbeault acknowledges funding by the Deutsche Forschungsgemeinschaft (DFG, German Research Foundation) - Project number 442047500 through the Collaborative Research Center ``Sparsity and Singular Structures'' (SFB 1481).

\appendix
\section{Metric graphs}
\label{sec:metric-graphs}
Here we recall several notions about graphs that are necessary in our developments. The presentation is based on the book \cite{BK2013}, which provides a comprehensive introduction to quantum graphs, and on the paper \cite{AB2018}, which develops finite element discretizations of elliptic operators in quantum graphs. We sometimes narrow down the generality of certain notions for the purposes of the present paper.

A \emph{combinatorial graph} $\G = (\V, \E)$ is a collection of a finite number of vertices $\V$ and of edges $\E\subset \V\times \V$ connecting pairs of vertices.
We restrict our attention to \emph{undirected} graphs where no orientation is assigned to the edges, and denote $|\V|$ and $|\E|$ the number of vertices and edges, respectively.

We will work with \emph{connected} graphs, where any two vertices $\vv, \ww\in\V$ are connected by at least one path $(\vv, \vv_1), (\vv_1, \vv_2),\dots, (\vv_k, \ww)$ made by consecutive adjacent edges in $\E$. A connected graph becomes a \emph{metric graph} if we assign a length $\ell_\ee >0$ and a local coordinate $r_\ee(x)$, for $x\in [0,\ell_\ee]$, to each edge $\ee=(\vv,\ww)\in\E$, in such a way that $r_\ee(0)=\vv$ and $r_\ee(\ell_\ee)=\ww$.

In our case, the crossroads $\V$ are embedded in $\R^2$ through their geographical coordinates, and the streets $r_\ee([0,\ell_\ee])\subset \R^2$ are differentiable curves with no loops. However, as done very often, we redefine them as simple straight lines joining the two vertices. Regardless of the choice of the edge curves, the points $r$ in a metric graph $\G$ are thus not only its vertices but also all intermediate points on the edges as well, parametrized by the local coordinates $r_\ee$:
\[
\V\subsetneq \G=\bigcup_{\ee\in \E} r_\ee([0,\ell_\ee]).
\]

As the name suggests, any metric graph can be endowed with a natural metric as follows. The distance between two vertices $\vv,\,\ww\in \V$ is as usual defined as the length of the shortest path connecting them. This notion of distance between vertices is then extended in a natural way to any two points possibly lying on different edges, by further adding the local coordinates along these edges.
\newline

We may now introduce function spaces and linear differential operators on a metric graph~$\G$.
The space of continuous functions $\cC(\G)$ contains the functions $u:\G\to \R$ such that $u\circ r_\ee$ is continuous on $[0,\ell_\ee]$ for any edge $\ee\in \E$, which implies in particular the continuity of $u$ along any path in $\G$.
The space of square-integrable functions
\[
L^2(\G) = \bigoplus_{\ee\in \E} L^2(r_\ee([0,\ell_\ee]))
\]
is a Hilbert space when endowed with the inner product
\[
\<u, v \>_{L^2(\G)}
\coloneqq \int_\G u(r)v(r)\dr
=\sum_{\ee\in \E} \int_0^{\ell_\ee} u(r_{\ee}(x)) v(r_{\ee}(x))dx.
\]
Finally, the Sobolev space 
\[
H^1(\G) =  \cC(\G)\cap\bigoplus_{\ee\in \E} H^1(r_\ee([0,\ell_\ee]))
\]
is also a Hilbert space, for the norm
\[
\Vert u \Vert^2_{H^1(\G)} \coloneqq \int_\G u^2\dr + \int_\G \left( \frac{\dd u}{\dr} \right)^2 \dr = \sum_{\ee\in \E} \int_0^{\ell_\ee} u(r_{\ee}(x))^2 dx+ \int_\G\left( \frac{\dd (u\circ r_\ee)}{dx} \right)^2 dx.
\]

The restriction to $\cC(\G)$ in the definition of $H^1(\G)$ stems from the fact that functions in $H^1(r_\ee([0,\ell_\ee]))$ are continuous (because their domain is one dimensional), which automatically implies that functions in $H^1(\G)$ must be continuous also at the vertices. In the same vein, one has to impose restrictions on the derivatives of $u$, such as Newmann-Kirchoff boundary conditions, for functions $u\in H^2(\G)$.

\bibliographystyle{unsrt}
\bibliography{literature}

\end{document}